\documentclass[runningheads]{llncs}
\usepackage{graphicx}
\usepackage{makecell}
\usepackage[square,sort,comma,numbers]{natbib}
\usepackage{ragged2e}
\begin{document}
\title{ShufText: A Simple Black Box Approach to Evaluate the Fragility of Text Classification Models \thanks{Supported by L3Cube Pune.}}
\titlerunning{ShufText: Evaluating Fragility of Text Classification Models}
%
\author{Rutuja Taware\inst{1} \and 
Shraddha Varat\inst{1} \and 
Gaurav Salunke\inst{1} \and 
Chaitanya Gawande\inst{1} \and 
Geetanjali Kale\inst{1} \and 
Rahul Khengare\inst{1} \and
Raviraj Joshi\inst{2} }
\authorrunning{R. Taware et al.}
%
\institute{Pune Institute of Computer Technology, Pune, Maharashtra, India \email{\{rutujam77,shraddhavarat77,gssalunke1999,cgawande12\}@gmail.com , gvkale@pict.edu, rahulk1306@gmail.com }  \and
Indian Institute of Technology Madras, Chennai,  Tamilnadu, India
\email{ravirajoshi@gmail.com}}
%
\maketitle              
\begin{abstract}
Text classification is the most basic natural language processing task. It has a wide range of applications ranging from sentiment analysis to topic classification. Recently, deep learning approaches based on CNN, LSTM, and Transformers have been the de facto approach for text classification. 
In this work, we highlight a common issue associated with these approaches. We show that these systems are over-reliant on the important words present in the text that are useful for classification. With limited training data and discriminative training strategy, these approaches tend to ignore the semantic meaning of the sentence and rather just focus on keywords or important n-grams. We propose a simple black box technique ShutText to present the shortcomings of the model and identify the over-reliance of the model on keywords. This involves randomly shuffling the words in a sentence and evaluating the classification accuracy. We see that on common text classification datasets there is very little effect of shuffling and with high probability these models predict the original class. We also evaluate the effect of language model pretraining on these models and try to answer questions around model robustness to out of domain sentences. We show that simple models based on CNN or LSTM as well as complex models like BERT are questionable in terms of their syntactic and semantic understanding.

\end{abstract}
\section{Introduction}
The domain of natural language processing involves multiple tasks such as text classification, language translation, text summarization, etc. With recent advances in deep neural networks (DNNs), neural network approaches are known to perform best on these tasks \cite{kowsari2019text}. Text classification is the most fundamental task which has a variety of applications. It forms the basis of sentiment analysis, document classification, and news categorization. It is also used for intent identification in chatbot systems \cite{liu2016attention}. These techniques have also been evaluated for domains like biomedical text \cite{rios2015convolutional} and non-English languages \cite{joshi2019deep,kulkarni2021experimental}.

\begin{figure}
\centering
\includegraphics[scale=0.2]{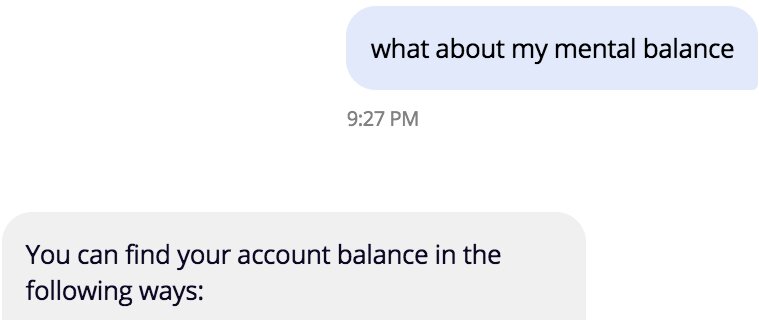}
\caption{Bank Chatbot}
\label{fig1}
\end{figure}

Deep learning approaches based on convolutional neural networks (CNNs) and recurrent neural networks (RNNs) are the most popular techniques used for text classification \cite{kim2014convolutional} \cite{zhou2015c}. Pre-trained networks like Bidirectional Encoder Representations from Transformers (BERT) based on transformer architecture perform better than networks trained from scratch \cite{paper8}. Our work is centered around text classification and these deep learning models. While these models have produced a state of the art results on text classification datasets it is difficult to interpret the classification results \cite{yuan2019interpreting}. There has been a lot of emphasis on model interpretation in the NLP community \cite{ribeiro2016should, wallace2019allennlp,feng2018pathologies,ribeiro2018semantically,conneau2018you,ribeiro2020beyond,deyoung2019eraser}. We would ideally expect these models to understand the text at the syntactic level and semantic level and arrive at a conclusion. However, recent works in the area of interpretable NLP has shown these models to be n-gram detectors \cite{jacovi2018understanding,li2016understanding}. These models lookout for specific n-grams to find the target class. For example in the case of sentiment analysis words like "happy", "sad", "great", "mediocre" play an important role in sentiment detection. Moreover, a typical sentiment analysis model would predict a neural sentence "I am reading the book Happy Days" as having a positive sentiment. This is mainly because of limited non-diverse training data and discriminative training such that the model tends to learn signals that are important for classification and ignores understanding the semantics of the sentence. In this work, we highlight this limitation of text classification models using a simple black-box approach. We show that these classification techniques are over-reliant on words that are important for classification decisions. It does not take into consideration the syntax or grammar and semantic meaning of the sentence. To motivate the problem let us look at two queries fired at a banking chatbot (refer Fig \ref{fig1}):
\begin{itemize}
\item What is my bank balance?
\item What about my mental balance?
\end{itemize}

Although the second query is not related to bank balance a typical chatbot still displays options to find one's account balance. This shows that the intent detection model which is also a text classification model was not able to understand the semantics of this out of domain query. As humans, we understand the meaning at the sentence level which these natural language understanding systems are unable to capture. There are real-world examples of Twitter and LinkedIn auto-reply bots misinterpreting the context \cite{AmazonCo61:online} and we believe it to be related to problems described in this work.

Moreover, studies in the past have discovered the vulnerabilities in DNNs by devising various adversarial attacks \cite{nguyen2015deep}. These attacks have shown that DNN can be easily fooled by making small perturbations in the input sample. These perturbations primarily focus on changing important keywords present in a sentence. Even though such models are deployed widely, they are still not robust enough to ensure high standards of security. Our work is not concerned with attacks on DNN but we aim to understand the ability of the model to comprehend sentence semantics.

In this work, we present a simple baseline technique - ShufText, which can be used to study the fragility of a model. The baseline acts as an indicator of the over-reliance of the model on specific keywords. We do so by creating a shuffled test set where the words in the test sentences are shuffled randomly but the labels still correspond to the original sentences. High accuracy on this test set would imply that the model has not learned the sentence structure or semantics and is relying on the key-words. Analyzing the prediction probabilities would provide further insights into the severity of the problem. Models pre-trained on large external corpus using language modeling objectives are also evaluated on this shuffled test set to check for their robustness. After observing that these models provide high accuracy on the shuffled test set, we perform a series of experiment to answer the following questions:

\begin{itemize}
\item Are these models capable of identifying shuffled sentences even when they are made part of training data and assigned a new label?
\item What is the effect of adding a generic class that represents out of domain sentences to the original train data?
\end{itemize}
The main contributions of this work are:
\begin{itemize}
\item We show that text classification models are heavily reliant on key-words or n-grams and ignore understanding the meaning of sentences.
\item We propose a simple black box technique- ShufText to quantize this shortcoming of the model.
\item Model performance after adding shuffled sentences and generic non-shuffled sentences to the training data is evaluated. The raw(non-pretrained) models are unable to distinguish between original and shuffled sentences even after adding it to the train set.
\end{itemize}

The paper is organized as follows. We present an overview of various existing works related to the generation of adversarial text samples in Section 2. The experimental setup is discussed in Section 3. The proposed ShufText approach is elaborated in Section 4. The data augmentation methods are presented in Section 5. Finally, the findings are summarized in Section 6.

\section{Related Work}
In this section, we review literature related to the fooling of DNNs using adversarial text samples. Although our work is not directly related to fooling we share the process of creating problematic text samples to validate the target task. 

Bin Liang et al. in \cite{paper3} generated adversarial samples by following two different types of attacks viz. white box and black box. In the white box attack, they used gradient computation to identify important text items, whereas, in the black box attack, they used a word occlusion technique to identify important keywords. Based on the identified keywords, three perturbation strategies viz. insertion wherein new text items were inserted,  modification wherein important text items were replaced with common misspellings, and removal was undertaken to generate final adversarial samples. The samples were then fed to character and word level DNNs and the results indicated that the models were susceptible to such attacks. 

Ji Gao et al. in \cite{paper4} proposed a black box technique that selected important keywords based on novel scoring functions. These keywords were then converted to unknown tokens by various transformation techniques and the resultant samples were fed to character level CNN and word-level LSTM models. The resultant samples were correctly classified by humans but the models gave wrong results with high confidence. This work also indicated that the generated adversarial samples were transferable i.e they can be used for fooling other models as well. Hence, when such samples were included in the trainset, adversarial accuracy increased for such models.

Javid Ebrahimi et al. in \cite{paper5} used gradient computation, a white box technique to identify important characters, and used beam search to identify the best combination of characters for substitution. These character level modifications were effective in fooling the networks and also generated robust models when such adversarial samples were included in the trainset. 

Utkarsh Desai et al. in \cite{paper6} have focused on the generation of a standard test set that can be used to expose the vulnerabilities of a model. This test set consists of real-world sentences made by adding random and targeted corruptions such as misspellings, text noise, etc in standard sentences. This paper further elaborates that the accuracy of standardized models decreases when tested on such samples.

Thus, the research work mentioned in this section greatly emphasizes the model reliance on keywords. Our work can also be viewed as a black box technique to study the fragility of a model.

\section{Experimental Setup}

\subsection{Datasets}
We use Stanford Sentiment Treebank (SST-2) \cite{paper1} and Text REtrieval Conference (TREC-6) \cite{paper2} datasets to evaluate the robustness of the models. These datasets were selected in order to cover binary and multi-class classification. The SST-2 dataset has positive and negative movie reviews with 6920 trainset samples and 1821 test set samples. The TREC-6 dataset consists of questions categorized into 6 classes: Abbreviation, Description and abstract concepts, Entities, Human beings, Locations, and Numeric values. This dataset has 5452 samples in the trainset and 500 samples in the test set. The data augmentation phase makes use of the Wikipedia dataset obtained from English Wikipedia articles. It was manually ensured that the selected Wiki articles did not overlap with the domain of SST and TREC datasets.

\subsection{Models}
In this work, we have performed experiments on four widely used text classification models. 
Our first classifier is a word-level CNN network that uses a single convolutional layer followed by a MaxPooling layer. Two linear layers are stacked upon the MaxPooling layer and a dropout layer is added after the convolutional layer and the first linear layer. We have used a uni-directional word-level LSTM network as the second classifier with two linear layers stacked on top of two LSTM layers and a dropout layer is added after the topmost LSTM layer and after the first linear layer.

The third classifier is the Universal Language Model Fine-tuning for Text Classification (ULMFiT) with an embedding layer of 300 hidden units, followed by three LSTM layers wherein the first two layers have 1150 hidden units and the last layer has 300 hidden units. Two linear layers are stacked upon the LSTM layers and dropout is added in every layer for regularization. For each layer, the learning rate increases linearly in the initial phase, and then it decays linearly generating a slanted triangular pattern \cite{paper7}. Our fourth classifier is a BERT large uncased model with 24 layers of bidirectional Transformer based encoder blocks wherein each layer has a total of 16 self-attention heads \cite{paper8}.

\section{ShufText}
The proposed ShufText approach involves creating a new test set by shuffling the words of original test sentences. The model is evaluated on this test set to measure the bias of the model towards relevant keywords. The sentences in the new test set are not meaningful and at the same time grammatically incorrect. The ground truth labels of these sentences are still the original labels. For example in the context of a banking application the shuffled sentence \textit{"bank balance my what is"} is still labeled as \textit{get\_bank\_balance}. If the trained model performs well on this dataset then it is an indicator of the limitation of the model to capture semantic meaning. Although it is obvious that the closest thing model can predict here is \textit{get\_bank\_balance} but a high prediction score is not obvious. The probability of predictions helps us quantize the severity of the problem. Since the words in the text are randomly shuffled the approach is termed as ShufText.

The following examples show how the text samples look before and after shuffling:
\begin{enumerate}
  \item Original text: When did Hawaii become a state
  
Corresponding shuffled text: Hawaii state a When did become
  \item Original text: offers that rare combination of entertainment and education
  
Corresponding shuffled text: that entertainment education rare and offers of combination
\end{enumerate}

\subsection{Method}
For a given model and a dataset consisting of a train set and a test set, the following steps are followed:

\begin{enumerate}
    \item The model is trained on the available train set.
    \item The original test set is passed through the model and class labels are obtained.
    \item Correctly classified samples are selected from the test set and the words of each sample are shuffled randomly.
    \item Accuracy on this shuffled test set is evaluated by comparing them with the original labels. 
    \item The above accuracy is termed as a percentage of same prediction i.e the predicted label of the shuffled samples is exactly the same as that of its original unshuffled version. A high accuracy would indicate that model is over-reliant on keywords and ignores the syntactic and semantic structure of the sentence.
    \item The probability of predictions for the original test set and shuffled versions are recorded. This helped us analyze how confident a model is while making wrong predictions.
\end{enumerate}

\begin{figure}
\includegraphics[width=\textwidth]{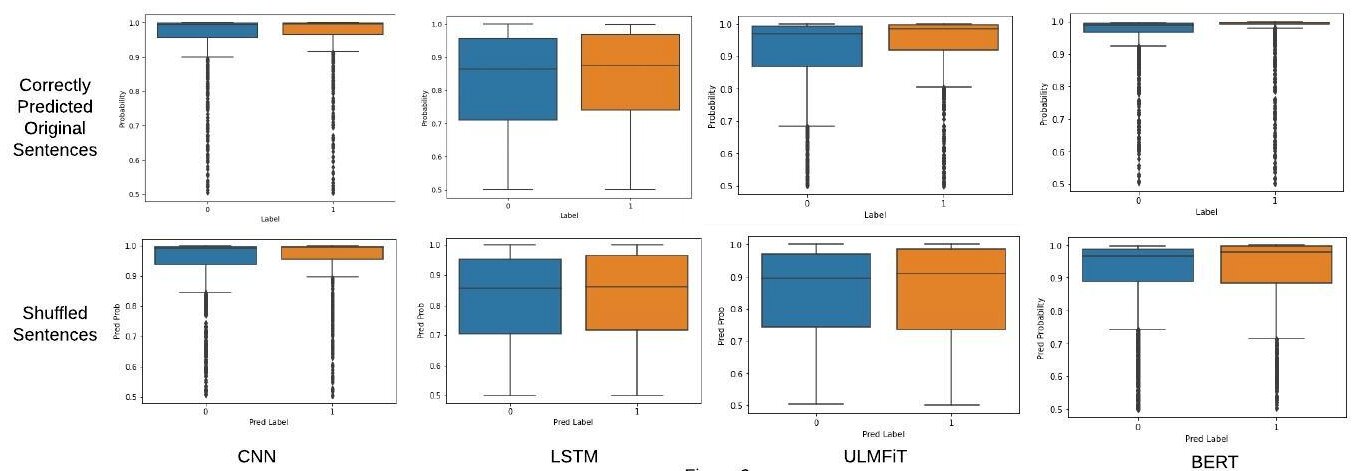}
\caption{ShufText SST} \label{fig2}
\end{figure}

A box plot is drawn between the class labels predicted by the model and the probabilities associated with them. These plots are drawn for the newly created shuffled test set as well as the original test set. In the case of the original test set, only correctly classified sentences are chosen for drawing the plots. This is done to ensure a fair comparison of the probability scores.

The plots are used to analyze the predicted probabilities for each class. Higher median probabilities in the box plot indicate a highly confident model whereas higher width of the box plots indicates a less confident model. 

\subsection{Result}

\begin{figure}
\includegraphics[width=\textwidth]{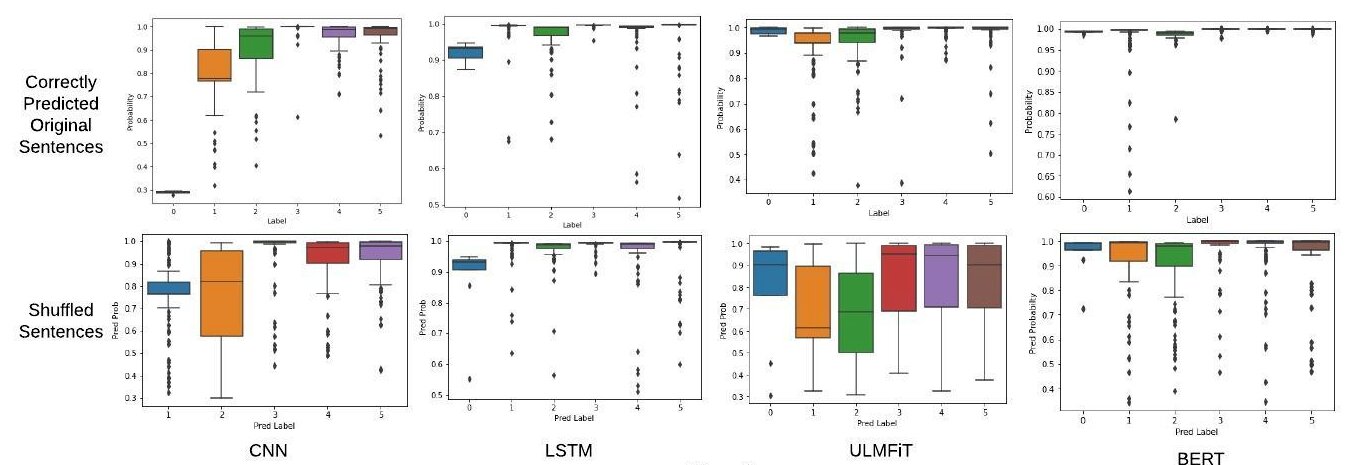}
\caption{ShufText TREC} \label{fig3}
\end{figure}

For simple models like CNN and LSTM, the plots of the correctly classified original test set samples and shuffled test set samples are very similar to each other. No significant changes can be observed in the median probability as well as the width of the boxes. Moreover, the percentage of the same prediction for these simple models is very high Table \ref{table1}. This indicates that shuffling the test set samples which were correctly classified by the models, does not have a major effect on their predictions. This shows that simple models are heavily dependent on keywords.

However, for BERT, the plots for shuffled sentences report a slight increase in the width of the boxes, refer Figures \ref{fig2}, \ref{fig3}. Additionally, the graph plots of shuffled sentences for the ULMFiT classifier indicate a major drop in its median probability and also a significant increase in its width, refer Figures \ref{fig2}, \ref{fig3}.  Moreover, the percentage of the same prediction for these pre-trained models is comparatively low, refer Table \ref{table1}. This shows that the models are not completely sure about the shuffled sentences which in turn indicates that they focus on the semantic connections between the keywords to some extent. 

However, for both the type of models i.e simple and pre-trained, the median probability for shuffled sentences is above 0.85 in the majority of cases. Therefore, even though the pre-trained models show a decrease in confidence while classifying shuffled sentences, it is evident that this decrease is not significant enough. This shows that even complex and highly accurate models like BERT are questionable as far as robustness is concerned.

\begin{table}
\begin{center}
\caption{ShufText Results}\label{table1}
\begin{tabular}{l c c c c }
\hline
       & \multicolumn{2}{c}{SST-2}                             & \multicolumn{2}{c}{TREC-6}                            \\ \hline
Model  & \makecell{Original \\ Test \\ Accuracy} & \makecell{Percentage \\ of Same \\ Prediction} & \makecell{Original \\ Test \\ Accuracy} & \makecell{Percentage \\ of Same \\ Prediction} \\ \hline
CNN    & 79.46                  & 93.29                         & 84.79                  & 85.84                         \\ 
LSTM   & 80.28                  & 92.20                         & 82.59                  & 94.18                         \\ 
ULMFiT & 85.83                  & 83.42                         & 93.00                  & 71.18                         \\ 
BERT   & 92.53                  & 83.02                         & 95.00                  & 66.52                         \\ \hline
\end{tabular}
\end{center}
\end{table}

\section{Data Augmentation Experiments}

\begin{figure}
\includegraphics[width=0.8\textwidth]{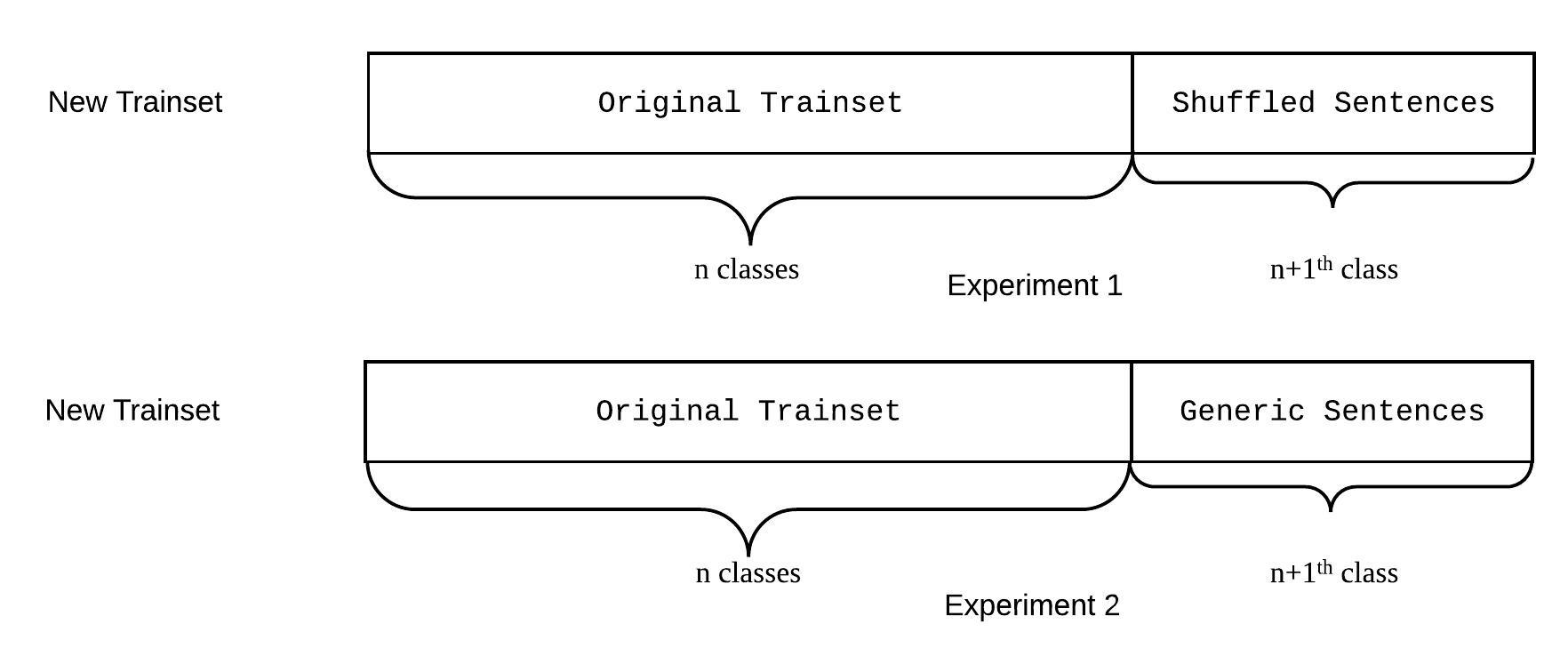}
\caption{Data Augmentation} \label{fig4}
\end{figure}

We have conducted a few data augmentation experiments and studied the effects that they have on the classification of standard sentences in the test set as well as their shuffled versions. The training data is augmented by adding shuffled and generic sentences. The idea is to see the effect of this addition on the test accuracy. The previous models were not capable of discriminating between shuffled and original sentences. So we experiment by explicitly adding shuffled sentences in training data. The addition of generic out of domain sentences is a more natural setup given natural language processing (NLP) applications. So, we also experiment with the addition of generic sentences instead of shuffled sentences. The exact process and results of adding shuffled sentences are described in Experiment 1 and the experiment related to generic sentences is described in the Experiment 2 section. It is ensured that the number of samples in the new class matches the number of samples in other existing classes so that all the classes have an equal ratio of sample size, refer Figure \ref{fig4}.

\subsection{ Experiment 1 - Addition of shuffled sentences}
In this experiment, the train set includes an additional class whose text samples are obtained by shuffling the sentences belonging to the original classes. The methodology and details of the experiment are outlined below.

\subsubsection{Method:}

\begin{figure}
\includegraphics[width=\textwidth]{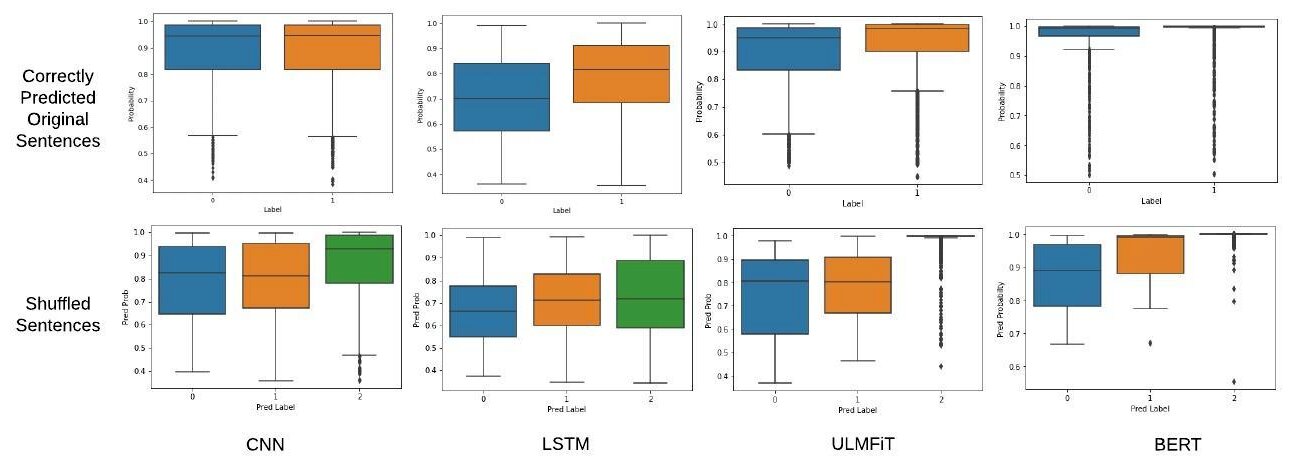}
\caption{Experiment 1 SST} \label{fig5}
\end{figure}

\begin{figure}
\includegraphics[width=\textwidth]{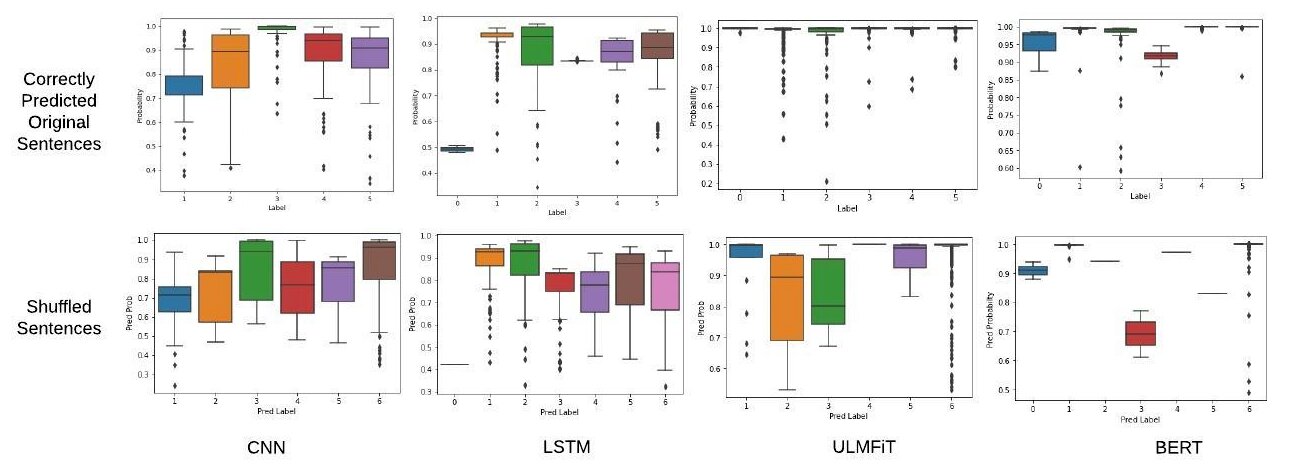}
\caption{Experiment 1 TREC} \label{fig6}
\end{figure}

For a given model and a dataset consisting of a train set and a test set, the following steps are followed:
\begin{enumerate}
    \item  $N$ random text samples present in the train set are shuffled where $N$ refers to the average number of samples present in each class.

    \item The shuffled samples obtained from the above step are appended to the train set and a new class label is assigned to these samples. The model is trained using the new train set.
    \item The test set is passed through the model and class labels are obtained.
    \item Correctly classified samples are selected from the test set and the words of each sample are shuffled randomly.
    \item The shuffled test set is passed through the model and the accuracy is evaluated.
    \item The probability of predictions for the original test set and shuffled versions are recorded. This helped us analyze how much a model is affected when shuffled sentences are added as a new class to the train set.
\end{enumerate}

\subsubsection{Result:}
The comparison of accuracy values between simple and pre-trained models shows that the CNN and LSTM classifiers yield very low accuracy values on the shuffled testset as compared to the pre-trained models, refer Table \ref{table2}. Hence, when shuffled sentences are made a part of the training set, the pre-trained models are better at distinguishing between shuffled and non-shuffled sentences. Although the task of distinguishing between shuffled and non-shuffled sentences is a trivial task, it becomes difficult for the simple networks, given original classes are maintained and the training data is limited.

Pre-trained models yield very high values for the overall accuracy of the test set, refer Table \ref{table2}. Hence, when the plots of original and shuffled samples are compared, it can be seen that these models yield a high probability for incorrect classification, but the fraction of these errors itself is very low, refer Figures \ref{fig5}, \ref{fig6}. Hence, the pre-trained models are able to distinguish between shuffled and non-shuffled sentences with high accuracy. These pre-trained models are capable of learning syntactic structures that simple models failed to recognize. This also highlights the importance of pre-training when training data is limited.

\begin{table}
\begin{center}
\caption{Experiment 1 Results}
\begin{tabular}{l c c c c c c}
\hline
       & \multicolumn{3}{c}{SST-2}                             & \multicolumn{3}{c}{TREC-6}                            \\ \hline
Model  & \makecell{Original \\Test \\Accuracy} & \makecell{Shuffled\\ Test \\Accuracy} & \makecell{Overall \\Test \\Accuracy} &  \makecell{Original \\Test \\Accuracy} & \makecell{Shuffled\\ Test \\Accuracy} & \makecell{Overall \\Test \\Accuracy}\\ \hline
CNN    & 71.77 & 61.28 & 67.39 & 85.0 & 72.23 & 79.13  \\ 
LSTM   &  68.09 & 33.06 & 53.90 & 81.59 & 28.43 & 57.70 \\ 
ULMFiT & 83.80 & 97.77 & 90.17 & 94.0 & 91.91 & 92.98 \\ 
BERT   & 91.87 & 97.96 & 94.79 & 96.2 & 94.59 & 95.41 \\ \hline

\end{tabular}
\label{table2}
\end{center}
\end{table}

\subsection{Experiment 2 - Addition of generic sentences}

In this experiment, the train set includes an additional class whose text samples contain generic sentences. These generic sentences are syntactically meaningful sentences and do not contain domain-specific keywords which will aid the model in the classification task. This technique is commonly used to detect out of domain sentences in a chat-bot like applications. We evaluate the effect of the addition of generic sentences on the model performance of simple and pre-trained models. The methodology and details of the experiment are outlined below.

\begin{figure}
\includegraphics[width=\textwidth]{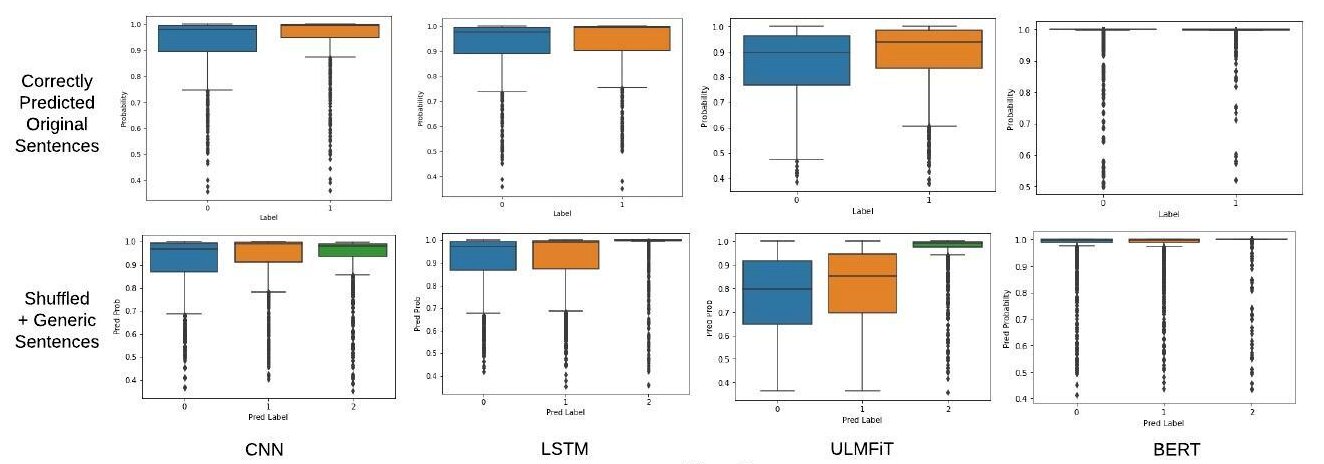}
\caption{Experiment 2 SST} \label{fig7}
\end{figure}

\begin{figure}
\includegraphics[width=\textwidth]{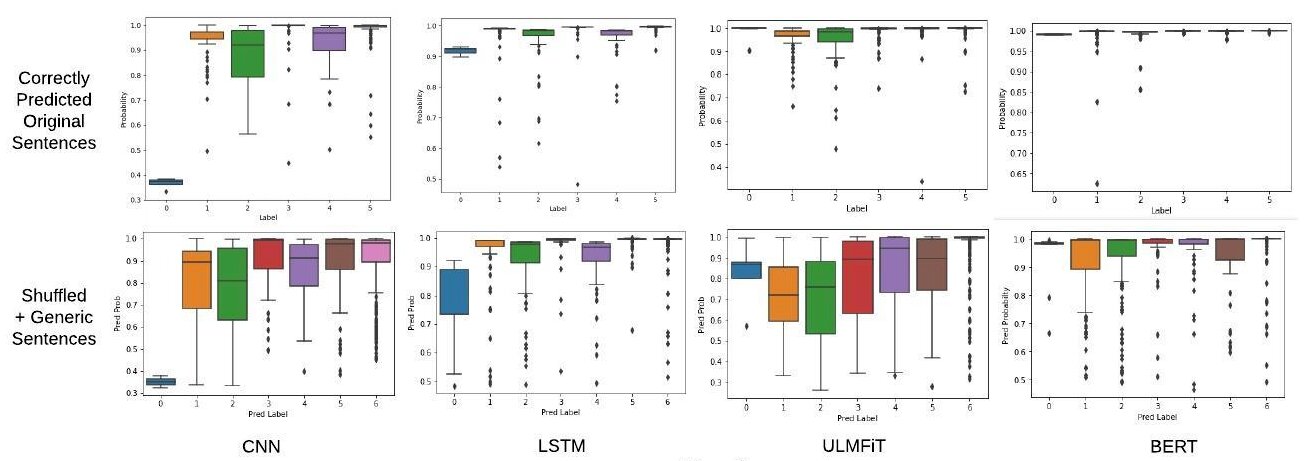}
\caption{Experiment 2 TREC} \label{fig8}
\end{figure}

\subsubsection{Method:}

For a given model and a dataset consisting of a train set and a test set, the following steps are followed:
\begin{enumerate}
    \item Generic sentences are obtained from the Wikipedia dataset such that the length of each sample lies in the interquartile range calculated from the lengths of the original samples present in the train set. $N$ generic text samples are chosen where $N$ refers to the average number of samples present in each class of the train set.
    \item The generic samples obtained from the above step are appended to the train set and a new class label is assigned to these samples. The model is trained using the new train set.
    \item The test set is passed through the model and class labels are obtained.
    \item Correctly classified samples are selected from the test set and the words of each sample are shuffled randomly.
    \item Generic sentences are obtained from the Wikipedia dataset such that the length of each sample lies in the interquartile range calculated from the lengths of the original samples present in the test set. These samples are different from those obtained in step 1.
    \item The samples obtained from steps 4 and 5 are combined to generate a new test set.
    \item The new test set is passed through the model and the accuracy is evaluated.
    \item The probability of predictions for the original test set and the new test set is recorded. This helped us analyze how much a model is affected when generic sentences are added as a new class to the train set. 
\end{enumerate}

\begin{table}
\begin{center}
\caption{Experiment 2 Results}
\begin{tabular}{l c c c c c c}
\hline
       & \multicolumn{3}{c}{SST-2}                             & \multicolumn{3}{c}{TREC-6}                            \\ \hline
Model  & \makecell{Original \\Test \\Accuracy} & \makecell{Generic\\ Sentence \\Accuracy} & \makecell{Percentage \\of same \\Prediction} &  \makecell{Original \\Test \\Accuracy} & \makecell{Generic\\ Sentence \\Accuracy} & \makecell{Percentage \\of same \\Prediction} \\ \hline
CNN    & 78.03 & 86.58   & 94.29 & 86.19    & 74.00   & 87.70  \\ 
LSTM   & 77.48   & 92.35    & 92.06      & 82.40    & 86.80 & 72.57    \\ 
ULMFiT & 82.75                  & 98.23                                                               & 82.81                                                                   & 94.20                  & 98.20                                                                 & 69.21    \\ 
BERT   & 90.82                  & 99.17                                                               & 83.91                                                                   & 96.40                  & 98.00                                                                 & 86.64    \\ \hline

\end{tabular}
\label{table3}
\end{center}
\end{table}
\subsubsection{Result:}
When generic sentences are added to the train set, all models yield a very high probability while classifying the original test set. Moreover, the accuracy on the original sentences of all the models does not show a major deflection when compared to ShufText, refer Tables \ref{table1}, \ref{table3}. Hence, it can be concluded that the addition of generic sentences does not hamper the confidence of simple and pre-trained models while classifying original sentences. Both types of models are equally capable of distinguishing between original sentences and out of domain generic sentences. This behavior was not exhibited for shuffled sentences by simple models.

For all the models, one can observe that the plots for shuffled test set obtained from Strategy 2 and ShufText are very similar to each other, refer Figures \ref{fig2}, \ref{fig3}, \ref{fig7}, \ref{fig8}. Hence it can be concluded that the addition of generic sentences to the trainset does not affect the models’ performance on the original as well as shuffled sentences. This can again be attributed to the difference in keyword distribution seen in original sentences and generic sentences.

\section{Conclusion}
The NLP models are not very reliable when they are deployed and fed "real world" inputs. These models are heavily reliant on important keywords or n-grams relevant to the classification task. ShufText provides a simple approach to quantify how reliant a model is on keywords and how much the model understands the semantic or syntactic layout of the input samples. We have also performed two data augmentation experiments and analyzed model behavior in each case. The first experiment involves the addition of shuffled samples to the training data and can be used to quantify the effect pre-training has on models. The inclusion of generic sentences to the train set as an additional class has been commonly practiced in text classification tasks. The generic out of the domain sentences were added to the train set in the second experiment. We show that pre-trained models are effective in distinguishing between original and shuffled sentences whereas simple models fail to do so. Both the models are effective in distinguishing between original sentences and out of domain generic sentences.

\section*{Acknowledgements} This work was done under the L3Cube Pune mentorship program. We would like to express our gratitude towards our mentors at L3Cube for their continuous support and encouragement.

\bibliographystyle{splncs04}
\bibliography{main}
\end{document}